\documentclass[letterpaper]{article} 
\usepackage{aaai2026} 
\usepackage{times} 
\usepackage{helvet} 
\usepackage{courier} 
\usepackage[hyphens]{url} 
\usepackage{graphicx} 
\usepackage{natbib} 
\usepackage{caption} 
\usepackage{algorithm}
\usepackage{algorithmic}
\usepackage{multirow}
\usepackage{booktabs}

\usepackage{newfloat}
\usepackage{listings}
\DeclareCaptionStyle{ruled}{labelfont=normalfont,labelsep=colon,strut=off} 
\floatstyle{ruled}
\newfloat{listing}{tb}{lst}{}
\floatname{listing}{Listing}
\usepackage{color,xcolor}

\title{Why Does Deepfake Detection Fail in the Real World? \\ A Comprehensive Study and Dataset}
\author{
Yuzhou Yang\textsuperscript{\rm 1},
Haoyue Wang\textsuperscript{\rm 1},
Chunyang Li\textsuperscript{\rm 1},
Yiyu Qian\textsuperscript{\rm 2}, \\
Sheng Li\textsuperscript{\rm 1},
Yuchen Xie\textsuperscript{\rm 3},
Zhenxing Qian\textsuperscript{\rm 1},
Xinpeng Zhang\textsuperscript{\rm 1}
}
\affiliations{
 \textsuperscript{\rm 1}Fudan University \\
 \textsuperscript{\rm 2}University of Wisconsin - Madison\\
 \textsuperscript{\rm 3}Xi'an Jiaotong-Liverpool University\\

}

\usepackage{bibentry}

\begin{document}

\maketitle

\begin{abstract}
The rapid evolution of deepfake technologies, particularly with the emergence of advanced AIGC models, erodes digital trust and is even weaponized for fraud and harassment.
Although state-of-the-art detectors report high detection accuracy, they often fail in many real-world scenarios. 
This failure stems from a fundamental disconnect between controlled laboratory datasets and the ``in-the-wild'' characteristics of online content.
To dissect this failure, we conduct a comprehensive study of real-world deepfakes.
We present {DeepFake-MIX}, a comprehensive and challenging benchmark curated from 50,000+ video clips.
The dataset uniquely incorporates: 1) content from major social platforms (TikTok, YouTube, etc.) with native disturbances, and 2) a diverse ecosystem of 26 AIGC-driven forgery techniques.
Our evaluations reveal that current detectors, despite achieving near-perfect scores on existing datasets like FF++ (99.6\%). See their performance plummet heavily on DeepFake-MIX's collection of trending deepfakes from the real world, exposing critical vulnerabilities.
Moreover, we study the detectors' behavior with different factors, e.g., occlusions, lighting, head positions and motions, etc.
Our findings may provide the research community with a crucial diagnostic framework and a public dataset to build and validate the next generation of deepfake detectors that are robust to real-world challenges, which is a critical step towards mitigating the adverse social impacts of malicious AI-generated content.
\end{abstract}


\section{Introduction}

The advent of Artificial Intelligence Generated Content (AIGC) has ushered in an era where hyper-realistic deepfakes can be synthesized with unprecedented ease.
From personalized entertainment to malicious disinformation campaigns, advanced forgery techniques like HiFiFace~\cite{wang2021hififace} and Janus~\cite{wu2024janus} now enable seamless face swapping, attribute manipulation, and even full facial synthesis without traditional blending artifacts~\cite{SimSwap}. 
While democratizing creativity, this technological leap poses critical challenges: over 95\% of deepfake videos detected by Meta in 2023 were weaponized for fraud or political manipulation~\cite{yan2024df40}, eroding digital trust at an alarming scale.

Current defenses rely heavily on deepfake detection models trained on controlled laboratory datasets, such as FaceForensics++ (FF++)~\cite{rossler2019faceforensics++}, Celeb-DF~\cite{li2020celebdf}, etc.
However, malicious attacks via deepfakes in the real world use forgery methods that are newer and more diverse, which are not included in these datasets.
Additionally, forged artifacts within media may be hidden by attackers using tools or weakened during transmission over the internet. 
Such operations range from basic occlusion to other visual effects.
Therefore, while detectors~\cite{NPR, nguyen2024laa} achieve high detection accuracy on these benchmark datasets, it may be a dangerous illusion of effectiveness.

Two fundamental limitations decouple most existing datasets from the adversarial situation in the real world.
The first limitation is the \textit{temporal obsolescence}, where 82\% of existing datasets (e.g., FF++~\cite{rossler2019faceforensics++}) are built on forgery techniques older than three years, failing to represent cutting-edge AIGC models, e.g., {StableDiffusionv3.5}~\cite{SD3.5}, {FLUX}~\cite{flux2024}, etc. 
Secondly, the \textit{realism naivety}.
Laboratory-crafted videos lack adequate consideration of complex internet disturbances.
Traditional data augmentation may be effective in defense resolution compression or blur.
However, detecting user-generated content posted on social medias is even harder due to prevalent disturbances that exist, e.g., occlusions by text, effects, filters, etc.
These limitations restrict the ``effective'' detectors that achieve 99\% accuracy on FF++ to plummet to 68.2\% when tested against real-world deepfakes.

\begin{figure*}[!t]
\centering
\includegraphics[width=1.0\textwidth]{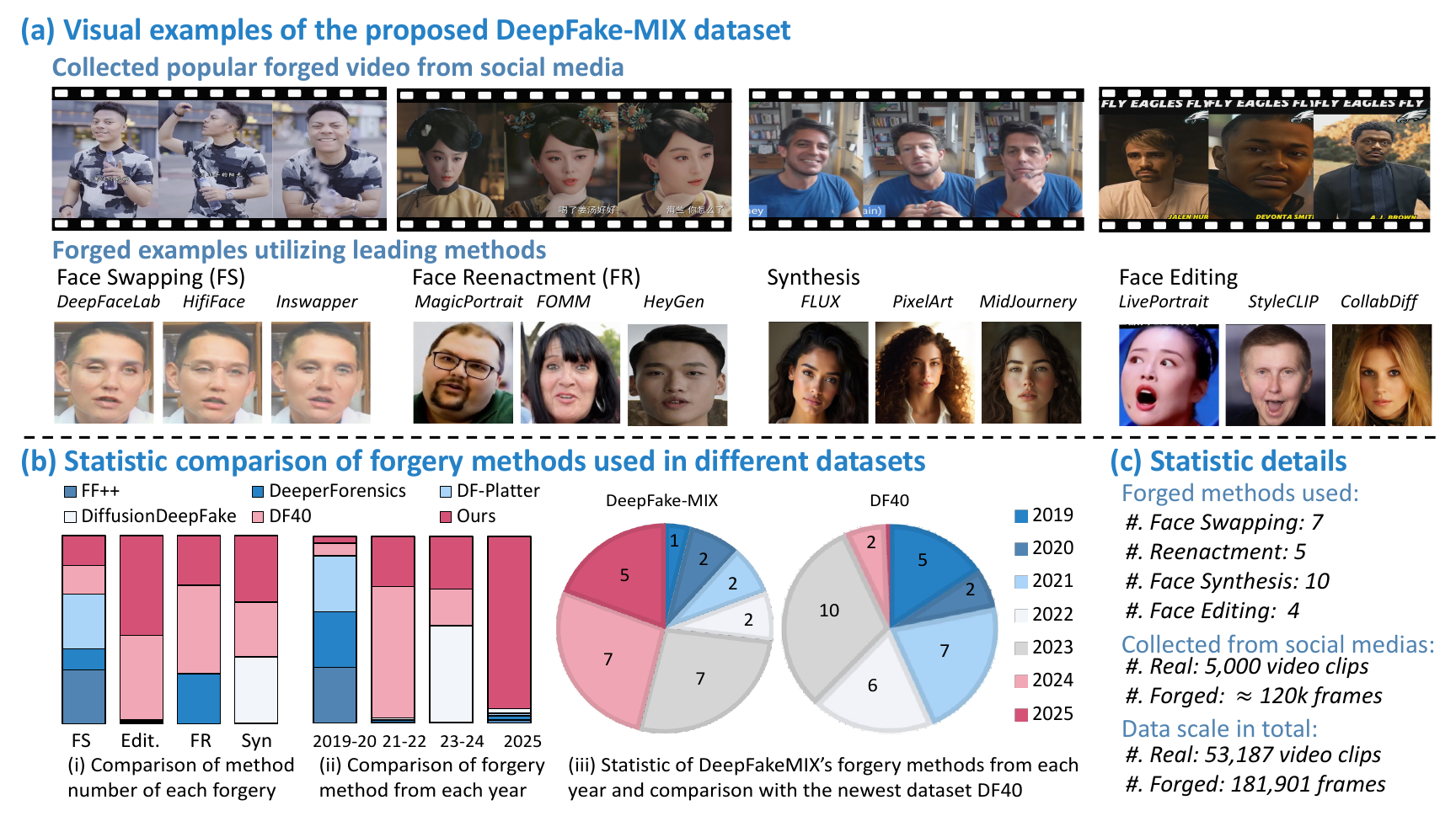} 
\caption{Overview of the proposed DeepFake-MIX dataset. The dataset consists of popular deepfake videos collected from social media and forged samples using diverse methods. We compare it with several commonly used deepfake datasets in terms of composition. We use \# denote ``the number of''.}
\label{fig:overview}
\end{figure*}

To move beyond laboratory conditions and systematically investigate why detectors fail,
we introduce DeepFake-MIX as the foundation for analysis.
The proposed dataset mirrors real-world deepfake evolution through two core innovations:
1) \textit{AIGC-Augmented Forgery Ecosystem}: integrating 26 state-of-the-art generation methods spanning face-swap, face synthesis, face reenactment, and attribute manipulation.
2) \textit{Realism Enhancement}: we collect over 5,000 video clips from prevalent platforms, preserving the native disturbances.
The overview of DeepFake-MIX is in Figure~\ref{fig:overview}.
Compared with other existing datasets, it demonstrates the diversity and updating of forgery methods.
Secondly, we propose an analytic framework for understanding deepfake creation and deployment in the wild, breaking it down into key stages where real-world factors introduce challenges.
Studied factors include illumination, head stances, emotions, and many post-processing techniques that were concluded from analyzing popular deepfakes on social media. The demonstration of the analytic framework is in Figure~\ref{fig:framework}.
Moreover, we present experiments with eight SoTA detectors under our analytic framework, quantify the failure of existing detectors, proving the severity of the problem.
Extensive evaluation of DeepFake-MIX reveals stark realities: models trained on FF++ achieve merely 80\% accuracy on DeepFake-MIX, while our dataset-trained variants show 10\% higher cross-domain accuracy.
These findings expose the fragility of current detection paradigms and establish DeepFake-MIX as an indispensable training foundation.

In conclusion, our key contributions include:
\begin{itemize}
\item We propose a large multi-source deepfake dataset (50k+ clips) with either preserved real-world perturbations or realism enhancements, while exhibiting strong AIGC-driven forgery diversity.
\item We comprehensively analyze the vulnerability of existing detectors under our framework. Providing some insight for training a robust detector.
\item We conduct extensive experiments using eight detectors, quantifying detection failures, and validating the effectiveness of the proposed dataset.
\end{itemize}

\section{Related Work}
\subsection{Deepfake Generation}

Deepfake techniques can typically be classified into four types, i.e., face swapping, reenactment, face synthesis, and face editing. 
The evolution of deepfake generation has progressed through several key phases. 
Early approaches primarily relied on autoencoder-based architectures for face swapping. 
The emergence of Generative Adversarial Networks (GANs)~\cite{Tero2020stylegan2} enables more realistic facial manipulations through adversarial training.
Modern techniques, e.g., HiFiFace~\cite{wang2021hififace} allow seamless face swapping with minimal artifacts. 
Recent developments in AIGC models~\cite{SD3.5, midjourney} introduce vivid face synthesis, significantly expanding the scope and realism of deepfake generation.

\subsection{Deepfake Detection}

Deepfake detection methods have evolved alongside the increasing sophistication of generation techniques. 
Early approaches focused on identifying specific artifacts or inconsistencies in manipulated content. ResNet~\cite{he2016resnet} and XceptionNet~\cite{chollet2017xception} demonstrate the effectiveness of deep learning architectures in capturing subtle manipulation patterns. 
Recent advances have introduced more sophisticated approaches, 
NPR~\cite{NPR} captures and characterizes the generalized structural artifacts stemming from up-sampling operations, causing generalized forgery artifacts for detection.
TruFor~\cite{guillaro2023trufor} utilizes a noise-sensitive fingerprint, combining low-level and high-level evidence, together with the confidence analysis, to provide more reliable decisions.  
UFD~\cite{ojha2023towards} utilizes features extracted by CLIP~\cite{clip}, integrating nearest neighbor and linear classification to achieve robust detection.
FreqNet~\cite{tan2024frequency} integrates a frequency domain learning module into the CNN, encouraging the detector to focus on high-frequency information.
LAA-Net~\cite{nguyen2024laa} combines heatmap-based and self-consistency attention strategies, utilizing a feature pyramid network to spread low-level features into the final feature for detection.

\subsection{Existing Deepfake Datasets}

Several benchmark datasets contribute to the deepfake detection.
FaceForensics++(FF++)~\cite{rossler2019faceforensics++} significantly expanded the previous datasets by incorporating multiple manipulation techniques and quality levels, while DeeperForensics~\cite{jiang2020deeperforensics1} introduced a larger collection of professionally crafted samples. 
DF-Platter~\cite{narayan2023df} contains 133,260 videos with diverse annotations, but only focuses on studying face swapping forgery.
DiffusionDeepFake~\cite{bhattacharyya2024diffusion} includes updating synthetic models, however, inconsiderate in other forgery types.
DF40~\cite{yan2024df40} contains 40 forgery methods. However, most methods are from before 2024, requiring an update.


The rapid advancement of AIGC technologies and the emergence of more sophisticated deepfake generation methods continue to challenge existing detection approaches.
Moreover, existing datasets primarily consist of laboratory-crafted samples with controlled conditions and often outdated manipulation techniques. 
These highlight the need for more comprehensive deepfake datasets.

\section{DeepFake-MIX Benchmark}

The DeepFake-MIX benchmark represents a comprehensive effort to bridge the gap between laboratory-crafted datasets and real-world deepfake scenarios. 
The ``MIX'' suffix represents the diverse nature of the dataset from construction to characteristic.

\begin{table}[!t]
\small
\centering
\begin{tabular}{rlllr} 
\hline
\textbf{ID} & \textbf{Method} & \textbf{Type} & \textbf{Format} & \textbf{Amount} \\ 
\hline
1 & Animate & Synthesis & Video & 150 \\
2 & MagicTime & Synthesis & Video & 150 \\
3 & SkyReels & Synthesis & Video & 50 \\
4 & Wan2.1 & Synthesis & Video & 150 \\
5 & Janus & Synthesis & Image & 150 \\
6 & FLUX & Synthesis & Image & 150 \\
7 & SD3.5 & Synthesis & Image & 150 \\
8 & MidJourney & Synthesis & Image & 1,000 \\
9 & PixelArt-$\alpha$ & Synthesis & Image & 2,000 \\
10 & StyleGAN-XL & Synthesis & Image & 2,000 \\
11 & UniFace & Face-swapping & Video & 1,000 \\
12 & HifiFace & Face-swapping & Video & 4,935 \\
13 & inswapper & Face-swapping & Video & 5,000 \\
14 & SimSwap & Face-swapping & Video & 5,000 \\
15 & E4S & Face-swapping & Video & 1,000 \\
16 & FaceDancer & Face-swapping & Video & 1,000 \\
17 & DeepFaceLab & Face-swapping & Video & 150 \\
18 & HeyGen & Reenactment & Video & 50 \\
19 & HyperReenact & Reenactment & Video & 1,000 \\
20 & FOMM & Reenactment & Video & 5,000 \\
21 & SadTalker & Reenactment & Video & 1,000 \\
22 & MagicPortrait & Reenactment & Video & 1,000 \\
23 & LivePortrait & Face Editing & Video & 5,000 \\
24 & CollabDiff & Face Editing & Image & 2,000 \\
25 & StyleCLIP & Face Editing & Image & 2,000 \\
26 & StarGANv2 & Face Editing & Image & 2,000 \\
\hline
\end{tabular}
\caption{Involved deepfake methods of the proposed dataset. We employ 25 distinct methods to create deepfake videos and images.}
\label{tab:dataset_stats}
\end{table}

\begin{figure*}[!t]
\centering
\includegraphics[width=1.0\textwidth]{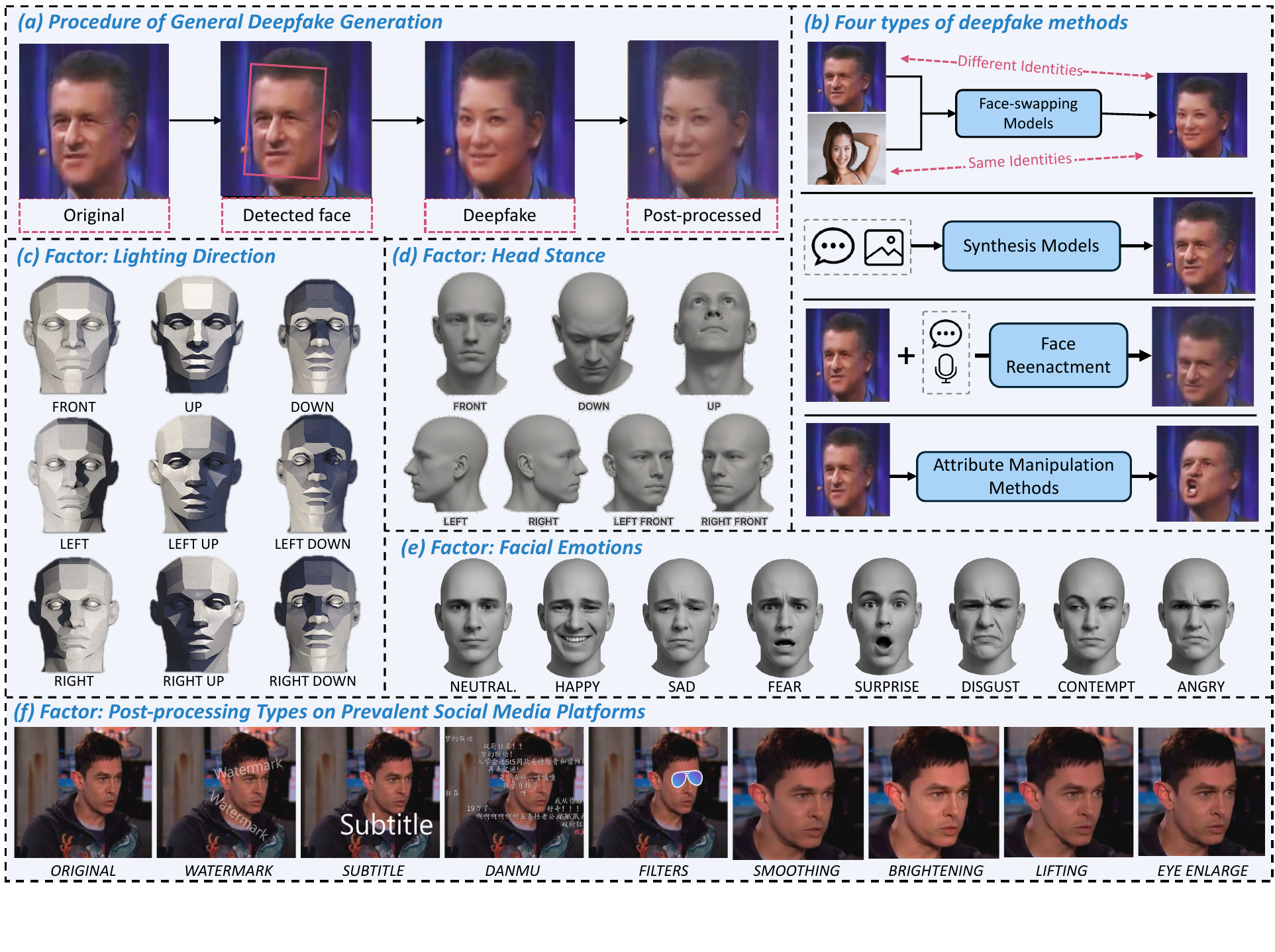} 
\caption{Demonstration of our analytic framework to study influential factors that impact deepfake detection. Results of factor (b), (c), (d), and (e), are presented in Table~\ref{tab:factor_model}, \ref{tab:factor_lighting}, \ref{tab:factor_stance}, and \ref{tab:factor_emotion}. Results of factor (f) are presented in Table~\ref{tab:factor_occlusion} and \ref{tab:factor_retouching}.}
\label{fig:framework}
\end{figure*}

\subsection{Dataset Construction}
\label{section:dataset_construction}
The DeepFake-MIX benchmark is constructed through a tripartite data construction architecture. Our methodology systematically integrates social media captured data, enhanced open-source data, and synthetic pipelines.

\noindent\textbf{Reality Captured Data.}
We collected real data and fake data from five major platforms, including TikTok (\url{tiktok.com}), YouTube(\url{youtube.com}), Xiaohongshu (\url{https://www.xiaohongshu.com}), BiliBili (\url{bilibili.com}), and Kuaishou (\url{https://www.kuaishou.com}), ensuring a diverse representation of real-world content for testing. 
The collection process focused on capturing various scenarios, including news broadcasts, interviews, vlogs, and entertainment content, to reflect the diverse contexts where deepfakes might appear.
In Figure~\ref{fig:overview}, we showcase the collected forgeries.
Collected samples have over 120,000 frames that contain deepfakes.

\noindent\textbf{Synthetic Pipeline.}
For comprehensive coverage of modern deepfake techniques, we integrate 26 state-of-the-art forgery methods, categorized into four main types.
(1) \textit{Face-sawapping} replaces the original face with the face of another person. We employ seven techniques, including UniFace~\cite{xu2022uniface}, Hififace~\cite{wang2021hififace}, inswapper~\footnote{https://github.com/deepinsight/insightface}, SimSwap~\cite{SimSwap}, E4S~\cite{liu2023fine}, FaceDancer~\cite{rosberg2023facedancer}, and DeepFaceLab~\cite{perov2020deepfacelab}.
The swapping faces are sampled from the CelebA~\cite{liu2015faceattributes}.
(2) \textit{Synthesis} methods directly generate virtual faces. 
We employ Animate~\cite{guo2023animatediff}, Magictime~\cite{yuan2025magictime}, SkyReels-V1~\footnote{\url{github.com/SkyworkAI/SkyReels-V1}}, Janus~\cite{wu2024janus}, StableDiffusionV3.5(SD3.5)~\cite{SD3.5}, Wan2.1~\cite{wang2025},  FLUX~\cite{flux2024}, SD3.5~\cite{SD3.5}, MidJourney~\cite{midjourney}, PixelArt~\cite{chen2024pixart}, and StyleGAN-XL~\cite{sauer2022stylegan}.
(3) \textit{Face reenactment} manipulates a person's facial movements to match another person's actions, speech, or predefined sequence. 
We employ FOMM~\cite{Siarohin_2019_NeurIPS}, HeyGen~\cite{heygen}, HyperReenact~\cite{bounareli2023hyperreenact}, SadTalker~\cite{zhang2023sadtalker}, and MagicPortrait~\cite{wei2025magicportrait}.
(4) \textit{Face editing} alters specific facial attributes, e.g., age, gender, hair color, etc., while preserving the person's identity. 
We employ LivePortrait~\cite{guo2024liveportrait}, CollabDiff~\cite{huang2023collaborative}, StyleCLIP~\cite{patashnik2021styleclip}, StarGANv2~\cite{choi2020stargan}.
Above are listed in Table~\ref{tab:dataset_stats}.

\noindent\textbf{Foundation Datasets and Enhancement.}
We employ three foundational datasets for composition, including 2,000 videos from FF++~\cite{rossler2019faceforensics++}, 3,434 videos from DeeperForensics~\cite{jiang2020deeperforensics1}, and 2,760 videos from Celeb-DF~\cite{li2020celebdf}.
We summarize several disturbances from data collected from social media and simulate these disturbances upon the foundation datasets.
Simulation includes:
1) \textit{Watermarks}: semi-transparent texts or logos with opacity from 20\% to 60\% with random positioning. 
2) \textit{Subtitles} with different sizes.
3) \textit{Danmu}: scrolling text with 60\% to 100\% opacity, which randomly obscures the whole video.
4) \textit{Decorative filters}: emojis and stickers with random locations and sizes.
5) \textit{Retouching}:  including skin smoothing, face brightening, face lifting, and eye enlargement, utilizing public tools.
We showcase examples of different enhancements in Figure~\ref{fig:framework}.


\subsection{Data Characteristics and Composition}

The DeepFake-MIX dataset is a comprehensive collection of synthetic media designed to reflect diverse forgery techniques and real-world variations, including 53,817 positive samples and 181,901 negative (forgery) samples. 

\noindent\textbf{Data Details.}
As detailed in Table~\ref{tab:dataset_stats}, 
1) Face-swapping methods contribute 14,935 high-resolution videos (up to 4K) with variable frame rates (20–120 FPS) and extended durations (up to 675 seconds), simulating realistic identity replacement scenarios.
2) Synthesis content includes video and image forgeries from novel methods. 
These entries emphasize high visual fidelity, though videos' durations are shorter (up to 5 seconds).
All content is generated under controlled prompts, 
in Figure~\ref{fig:overview} we exhibit results of the same prompt from three different models.
3) Face reenactment includes over 8,000 video clips, with a minimum resolution of 256$\times$256, and 
4) face editing includes 5,000 forged footages with resolution up to 4K, fps up to 120, and duration up to 1,700 seconds. It also includes over 6,000 images.
5) Social media collected videos have FPS ranges from 15 to 60, and resolution ranges from 480p to 1080p. 
The dataset’s scale and multi-technique forgeries enable robust evaluation of detection frameworks against evolving deepfake threats. 
This composition ensures coverage of both technical extremes and practical challenges, aligning with real-world deployment scenarios.

\noindent\textbf{Comparisons with other datasets.}
DeepFake-MIX demonstrates higher diversity in synthesis and incorporates more advanced forgery techniques compared to existing benchmarks. 
We compares with DiffusionDeepFake~\cite{bhattacharyya2024diffusion}, DF-Platter~\cite{narayan2023df}, DeeperForensics~\cite{jiang2020deeperforensics1},  FF++~\cite{rossler2019faceforensics++} and DF40~\cite{yan2024df40}. 
Our analysis examines both the quantity of techniques across forgery categories and the temporal progression of methods integrated into these datasets.
As shown in Figure~\ref{fig:overview} (b), DeepFake-MIX achieves unparalleled methodological diversity, supporting a broader spectrum of forgery types than any predecessor.
Furthermore, it uniquely incorporates state-of-the-art methods developed between 2021 and 2025, including diffusion-based architectures, GAN frameworks, and some commercial tools, ensuring alignment with evolving attack paradigms. 
This comprehensive coverage of cutting-edge techniques and temporal relevance positions DeepFake-MIX as a critical benchmark for evaluating detection systems against emerging sophisticated threats.

\section{Analysis of Influential Factors}

The general procedure of the deepfake creation can be summarized into three steps: face localization, deepfake generation, and post-processing.
In this section, we propose a framework for understanding the factors within the deepfake creation process that most influence detection.

\subsection{Factor I: Generation Models}
\label{section:factor1}
Deepfake images are synthesized by modifying localized facial regions using a series of generative models. 
As illustrated in Figure~\ref{fig:framework}(b), conventional deepfake techniques can be categorized into four primary types. 
Since different generation models inherently influence detection performance, we further delineate sub-categories for each type.
(1) \textbf{Face-swapping}: this paper selects four representative face-swapping methods. 
SimSwap~\cite{SimSwap} and
FaceDancer~\cite{rosberg2023facedancer} generate forged face and background for swapping.
E4S~\cite{liu2023fine} separates the texture of facial components and alters the identity by fine-grained changing.
DeepFaceLab~\cite{perov2020deepfacelab} disentangles the face area from the rest by localization, then only generates the face area for swapping.
(2) \textbf{Face Synthesis}: this technique produces ``AIGC faces''. 
With the rapid development of generation models, this technique has achieved notable improvement recently.
This paper selects representative GAN-based~\cite{sauer2022stylegan}, Diffusion-based ~\cite{chen2024pixart}, and a commercial application MidJounery~\cite{midjourney} for comparison.
(3) \textbf{Face Reenactment}: this technique modifies source faces, imitating the actions or expressions.
This paper studies different methods that are driven by audio, image, and text, i.e., SadTalker~\cite{zhang2023sadtalker}, HyperReenact~\cite{bounareli2023hyperreenact}, and HeyGen~\cite{heygen}.
(4) \textbf{Attribute Manipulation}: this technique aims to modify the facial attributes, e.g., emotions and age. 
This paper compares methods that utilize StyleGAN, diffusion, or others, i.e., StyleCLIP~\cite{patashnik2021styleclip}, CollabDiff~\cite{huang2023collaborative}, LivePortrait~\cite{guo2024liveportrait}.

\subsection{Factor II: Illumination Direction}
Inevitably, illumination direction causes inconsistent illumination conditions on the facial area. 
Low-intensity illumination conditions distort facial features during the localization phase. 
Overexposure washes out texture details, while shadows create unnatural contours. 
These irregularities propagate through subsequent forgery steps, introducing inconsistencies in skin tone and reflectance patterns.
We study the influence of illumination directions as conditioned in Figure~\ref{fig:framework}~(c), which significantly affect different areas on faces. 

\subsection{Factor III: Head Stances}
The orientation and angle of the head significantly impact facial landmark detection accuracy. Frontal faces are typically localized with high precision, while extreme poses (e.g., profile views, tilts) challenge alignment algorithms, creating artifacts or incomplete forgeries that leave detectable traces.

\subsection{Factor IV: Emotions}
Most models fail to replicate subtle muscle movements (e.g., brief eyebrow raises), creating temporal inconsistencies in videos. Generated emotions frequently lack correlated physiological signals like realistic pupil dilation or skin flushing patterns. 
We study the impact of eight common emotions that were defined in previous work~\cite{jiang2020deeperforensics1}.

\subsection{Factor V: Realistic Post-processing}
We study the impact of prevalent post-processing operations from social media platforms.
Those operations can be categorized into the following two types.
(1)~\textbf{Occlusion-based.}
Objects obstructing facial regions force models to infer hidden features. 
This extrapolation often results in implausible reconstructions around occluded areas (e.g., blurred eyes or distorted facial boundaries), yielding artifacts that impact detection methods.
This paper includes (i)~\textit{watermark}, (ii)~\textit{subtitle}, (iii)~\textit{danmu}, and (iv)~\textit{decorative filters} for study.
(2)~\textbf{Retouching-based.}
Retouching-based operations edit the image's color, brightness, regional features, etc. 
This paper examines four common retouching techniques on social media, including skin smoothing, face brightening, face lifting (affecting the contour of faces), and eye enlargement.

\begin{table*}[t]
\small
\centering
\setlength{\tabcolsep}{1.2mm}
\begin{tabular}{l|rrrrrrrrrrrrrrrrrr}
\hline
\multirow{2}{*}{\textbf{Method}} & \multicolumn{2}{c}{\textbf{Uniform}} & \multicolumn{2}{c}{\textbf{Up}} & \multicolumn{2}{c}{\textbf{RightUp}} & \multicolumn{2}{c}{\textbf{Right}} & \multicolumn{2}{c}{\textbf{RightDown}} & \multicolumn{2}{c}{\textbf{Down}} & \multicolumn{2}{c}{\textbf{LeftDown}} & \multicolumn{2}{c}{\textbf{Left}} & \multicolumn{2}{c}{\textbf{LeftUp}} \\
\cline{2-19}
& ACC & F1 & ACC & F1 & ACC & F1 & ACC & F1 & ACC & F1 & ACC & F1 & ACC & F1 & ACC & F1 & ACC & F1 \\
\hline
NPR & 66.4 & 62.0 & 56.7 & 72.4 & 67.4 & 80.5 & 74.3 & 85.3 & 87.6 & 93.3 & 76.0 & 86.3 & 66.7 & 80.2 & 91.0 & 95.4 & 61.2  & 66.1 \\
Freq-Net & 75.1 & 70.8 & 70.0 & 63.8 & 66.8 & 40.9 & 75.2 & 74.3 & 77.3 & 73.6 & 76.2 & 75.0 & 72.2 & 62.8 & 79.3 & 72.8 & 68.2 & 69.7 \\
LAA-Net & 66.1 & 75.7 & 62.7 & 72.0 & 73.1 & 82.5 & 70.9 & 81.6 & 63.8 & 76.4 & 60.3 & 69.7 & 76.3 & 84.9 & 76.8 & 85.5 & 77.6 & 85.2 \\
\hline
\end{tabular}
\caption{Results of \textbf{Factor II: Illumination Direction}. The table reports accuracy (ACC) and F1 score.}
\label{tab:factor_lighting}
\end{table*}

\begin{figure}[!t]
\centering
\includegraphics[width=1.0\linewidth]{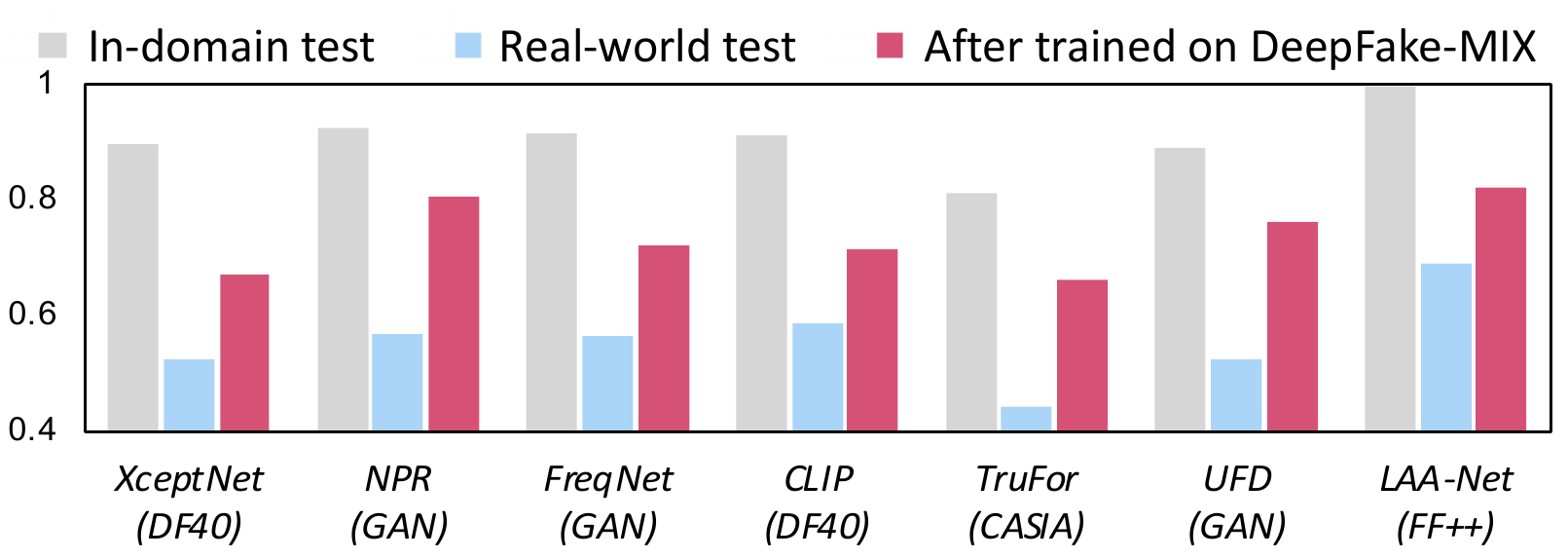} 
\caption{Comparisons of SoTA detectors' in-domain test results, pre-trained real-world test results, and real-world test results trained on DeepFake-MIX. e.g., UFD (GAN) indicates UFD trained on GAN-generated deepfakes, we report detection accuracy.}
\label{fig:real_world_test}
\end{figure}

\begin{table}[t]
\small
\centering
\setlength{\tabcolsep}{1mm}
\begin{tabular}{l|l|l|rr} 
\hline
\textbf{Type} & \textbf{Method} & \textbf{Description} & \textbf{ACC} & \textbf{F1} \\ 
\hline
\multirow{4}{*}{Swapping} & SimSwap & Re-generate & 81.8 & 90.0\\
 & FaceDancer & Re-generate & \textbf{79.3} & \textbf{88.5} \\
 & E4S & Fine-grained & 84.2 & 91.4 \\
 & DeepFaceLab & Disentangle & 93.7 & 96.8 \\ 
\hline
\multirow{3}{*}{Synthesis} & StyleGAN-XL & GAN-based & 95.6 & 97.7 \\
 & PixArt-$\alpha$ & Diffusion-based & 91.9 & 95.8 \\
 & MidJourney6 & Commercial & \textbf{91.3} & \textbf{95.4} \\ 
\hline
\multirow{4}{*}{Reenactment} & SadTalker & Audio-driven & \textbf{80.4} & \textbf{89.1}\\
 & HyperReenact & Image-driven & 95.7 & 97.8 \\
 & HeyGen & Text-driven & 92.8 & 96.3 \\ 
 & MagicPortrait & Video-driven & 88.4 & 91.5 \\ 
\hline
\multirow{3}{*}{Editing} & StyleCLIP & StyleGAN-based & 94.5 & 97.2 \\
 & CollabDiff & Diffusion-based & 92.6 & 96.2 \\
 & LivePortrait & Other & \textbf{91.4} & \textbf{94.8} \\
\hline
\end{tabular}
\caption{Results of different deepfake methods' impact. Detailed description is presented in Section~{Factor I}.}
\label{tab:factor_model}
\end{table}

\begin{table*}[t]
\small
\centering
\setlength{\tabcolsep}{1.8mm}
\begin{tabular}{l|rrrrrrrrrrrrrrrr}
\hline
\multirow{2}{*}{\textbf{Method}} & \multicolumn{2}{c}{\textbf{Neutral}} & \multicolumn{2}{c}{\textbf{Happy}} & \multicolumn{2}{c}{\textbf{Sad}} & \multicolumn{2}{c}{\textbf{Fear}} & \multicolumn{2}{c}{\textbf{Surprise}} & \multicolumn{2}{c}{\textbf{Disgust}} & \multicolumn{2}{c}{\textbf{Contempt}} & \multicolumn{2}{c}{\textbf{Angry}} \\
\cline{2-17}
& ACC & F1 & ACC & F1 & ACC & F1 & ACC & F1 & ACC & F1 & ACC & F1 & ACC & F1 & ACC & F1 \\
\hline
NPR & 78.6 & 64.5 & 76.6 & 61.5 & 75.3 & 60.1 & 74.8 & 59.0 & 75.0 & 60.0 & 74.3 & 58.6 & 74.1 & 58.2 & 74.3 & 60.6 \\
Freq-Net & 65.2 & 60.1 & 63.8 & 58.3 & 62.5 & 56.8 & 61.9 & 55.4 & 62.1 & 56.2 & 61.4 & 54.9 & 61.0 & 54.5 & 62.3 & 57.8 \\
LAA-Net & 67.3 & 76.7 & 61.4 & 71.0 & 64.2 & 75.5 & 65.2 & 75.2 & 63.3 & 72.5 & 60.7 & 71.5 & 60.0 & 71.0 & 64.2 & 74.9 \\
\hline
\end{tabular}
\caption{Results of \textbf{Factor IV: Emotions}.}
\label{tab:factor_emotion}
\end{table*}

\begin{table}[t]
\small
\centering
\setlength{\tabcolsep}{0.7mm}
\begin{tabular}{c|r|ccccccc}
\hline
\multicolumn{2}{c}{\textbf{Method}} & \textbf{Front} & \textbf{Left} & \textbf{L.Front} & \textbf{Right} & \textbf{R.Front} & \textbf{Down} & \textbf{Up} \\
\hline
\multirow{2}{*}{NPR} & ACC & 78.0 & 79.3 & 75.7 & 72.8 & 71.7 & 72.2 & 68.3 \\
& F1 & 64.2 & 78.1 & 86.1 & 81.4 & 83.6 & 75.3 & 80.1 \\
\hline
\multirow{2}{*}{Freq-Net} & ACC & 72.5 & 74.3 & 71.2 & 73.5 & 70.8 & 68.2 & 65.4 \\
& F1 & 70.3 & 82.5 & 81.0 & 84.2 & 80.1 & 75.3 & 73.2 \\
\hline
\multirow{2}{*}{LAA-Net} & ACC & 66.7 & 69.7 & 66.9 & 71.7 &71.9 & 66.0 & 61.7 \\
& F1 & 76.2 & 79.9 & 77.2 & 81.9 & 81.5 & 76.0 & 71.8 \\
\hline
\end{tabular}
\caption{Results of \textbf{Factor III: Head Stances}.}
\label{tab:factor_stance}
\end{table}

\begin{table}[ht]
\small
\centering
\begin{tabular}{lrrrrrr}
\hline
\textbf{Method} & \textbf{Base} & \textbf{WM} & \textbf{ST} & \textbf{DM} & \textbf{FL} & \textbf{Avg.} \\
\hline
\multicolumn{5}{l}{Trained on FF++} \\
\cline{1-2}
ResNet50 & 75.7 & {61.1} & {65.9} & {52.7} & {61.1} & {60.4} \\
XceptionNet & 80.3 & 58.6 & 62.4 & 48.7 & 57.7 & 56.8 \\
NPR & {84.9} & 56.1 & 58.9 & 44.6 & 54.3 & 54.0 \\
FreqNet & 71.0 & 55.0 & 51.8 & 34.9 & 55.2 & 51.9 \\
\hline
\multicolumn{5}{l}{Trained on CelebDF} \\
\cline{1-2}
ResNet50 & 89.2 & 60.4 & {49.5} & 46.3 & {58.0} & 56.5 \\
Xception & 89.9 & 60.5 & 48.2 & 48.7 & 57.5 & 53.8 \\
NPR & {90.6} & {60.7} & 46.9 & 51.2 & 57.1 & {56.7} \\
FreqNet & 85.7 & 54.1 & 46.7 & {68.6} & 56.7 & 56.4 \\
\hline
\multicolumn{5}{l}{Trained on DeeperForensic} \\
\cline{1-3}
ResNet50 & 99.8 & 65.8 & 72.6 & 82.0 & 67.3 & 69.3 \\
Xception & 99.4 & 66.0 & 73.4 & 81.2 & 67.3 & {72.0} \\
NPR & 99.0 & {66.2} & {74.8} & 80.5 & 67.2 & 69.4 \\
FreqNet & 98.9 & 66.1 & 72.1 & 78.9 & 66.8 & 68.7 \\
\hline
\end{tabular}
\caption{Results of testing \textbf{Factor V's occlusion-based processing}. Base denotes testing on the original testing data. WM (watermark), ST (subtitle), DM (danmu), FL (filters), Avg (average).}
\label{tab:factor_occlusion}
\end{table}

\begin{table}[ht]
\small
\centering
\begin{tabular}{l|l|rrrr} 
\hline
\textbf{Metric} & \textbf{Intensity} & \textbf{SM} & \textbf{FB} & \textbf{FL} & \textbf{EL} \\ 
\hline
\multirow{3}{*}{FN Rate} & None & 27.1 & 27.1 & 27.1 & 27.1 \\
 & Slight & 27.6 & 27.8 & 27.3 & 31.4 \\
 & Heavy & \textbf{39.3} & 28.6 & {31.6} & \textbf{33.6} \\ 
\hline
\multirow{3}{*}{FP Rate} & None & 21.7 & 21.7 & 21.7 & 21.7 \\
 & Slight & \textbf{33.3} & \textbf{30.9} & 25.6 & 20.5 \\
 & Heavy & 20.5 & \textbf{35.0} & \textbf{26.9}& 19.5 \\
\hline
\end{tabular}
\caption{Results of \textbf{Factor V's retouching-based processing}. The false negative (FN) rate and false positive (FP) rate of ResNet50's detection results on each retouching type with different intensities. SM (skin smoothing), FB (face brightening), FL (face lifting), EL (eye enlarging). Significant results are emboldened.}
\label{tab:factor_retouching}
\end{table}

\section{Experimental Results}

\subsection{Validation of DeepFake-MIX}

Eight SoTA detectors are selected for validation, including ResNet50~\cite{he2016resnet}, XceptionNet~\cite{chollet2017xception}, NPR~\cite{NPR}, FreqNet~\cite{tan2024frequency}, CLIP~\cite{clip}, TruFor~\cite{guillaro2023trufor}, UFD~\cite{ojha2023towards}, and LAA-Net~\cite{nguyen2024laa}.
We conduct experiments under two conditions to validate the effectiveness of DeepFakeMIX. The results are presented in Figure~\ref{fig:real_world_test}.

\noindent\textbf{Firstly, we test pre-trained models on real-world data.}
We compare the results of testing different pretrained methods on DeepFake-MIX's social media collected data. 
All methods exhibit performance drop compared to their testing on benchmarks, LAA-Net significantly plummeting from 99\% to 69\%.
Even methods trained on DF40 suffer a great decline. 
This suggests that laboratory strength doesn't necessarily equate to effectiveness in real-world applications.

\noindent\textbf{Then, we train detectors using DeepFake-MIX, then test on the real-world data.} 
All detectors trained on DeepFake-MIX generally exhibit strong generalization to real-world data. 
This suggests that the diversity and continuous updates in DeepFake-MIX contribute to training more robust and effective deepfake detectors.

\subsection{Analysis of Factors}

In this section, we implement SoTA detectors to generally evaluate different factors' impact on detection.
Factors II to IV's true data are selected from the foundation dataset DeeperForensics~\cite{jiang2020deeperforensics1}. We apply HifiFace, inswapper, and SimSwap on them randomly for study.

\noindent\textbf{Impact of different types of forgery methods is shown in Table~\ref{tab:factor_model}.} 
Among the four face-swapping techniques, re-generate methods—which synthesize both the forged face and background as a cohesive image—are the most challenging to detect. Fine-grained forgery methods present moderate difficulty, while techniques that manipulate only the facial region in isolation are the easiest to identify.
For face synthesis, diffusion-based forgery outperforms GAN-based approaches, with MidJourney6 being the most challenging to detect.
In face reenactment driven by different modalities, audio-driven methods are the hardest to detect, likely because their minimal alteration requirements leave fewer artifacts. Video-driven methods also pose significant challenges.
Among face editing methods, the exclusive LivePortrait technique is the most difficult to identify, while StyleGAN-based methods are the easiest.

\noindent\textbf{Impact of illumination direction is presented in Table~\ref{tab:factor_lighting}.}
Comparing three methods, uniform illumination does not improve detection accuracy.
Concentrated illumination (e.g., left- or right-sided lighting) can enhance detection by revealing subtle forgery artifacts, as the controlled brightness variations highlight inconsistencies.
However, extreme illumination conditions (e.g., top- or bottom-heavy lighting) severely degrade detection performance, likely due to exaggerated shadows or unnatural highlights that obscure facial details.
Additionally, different forgery detection methods exhibit varying sensitivities to specific illumination conditions.

\noindent\textbf{Impact of head stances is presented in Table~\ref{tab:factor_stance}.}
Unlike illumination conditions, head stance exhibits a less pronounced impact on detection performance for LAA-Net. 
This resilience may stem from modern forgery methods being trained on extensive datasets that inherently encompass diverse head orientations, thereby improving their robustness to pose variations~\cite{rossler2019faceforensics++}.
However, different training datasets of NPR and Freq-Net result in their less resilient ability.

\noindent\textbf{Impact of emotions is presented in Table~\ref{tab:factor_emotion}.}
Notably, all tested emotions introduce significant facial distortions, making forgery detection more challenging.
Among these, emotions that strongly affect eye movements (e.g., contempt and disgust) are the most disruptive to detection. Similarly, expressions involving wide mouth openings (e.g., happy and surprise) also substantially degrade detection accuracy.
In contrast, milder emotions like sadness and fear have minimal impact on detection performance.

\noindent\textbf{Impact of occlusion-based processing from social media is presented in Table~\ref{tab:factor_occlusion}.} 
Results show that all occlusion types reduce detection accuracy. 
Among them, those cause the most severe occlusion and have the strongest negative impact.
We randomly incorporate these occlusions into the training data, results show around 5\% accuracy improvement on real-world testing.

\noindent\textbf{Impact of retouching-based processing from social media is presented in Table~\ref{tab:factor_retouching}.} 
Key findings include: (i)~\textit{skin smoothing} makes forged samples slightly harder to detect under mild application but significantly increases misclassification of real images under heavy application. 
(ii)~\textit{face brightening} consistently reduces the detectability of forged samples across all intensities.
(iii)~\textit{face lifting} has a marginal effect on detection performance but may bias detectors toward classifying forged samples as real.
(iv)~\textit{eye enlargement}, when applied heavily, leads to a higher tendency of misclassifying real samples as forged.

\section{Discussion and Conclusion}

\textbf{Social Impact.}
Our study reveals that current detection systems—despite high laboratory performance—fail catastrophically against real-world deepfakes, leaving society vulnerable to fraud. 
The DeepFake-MIX dataset and analytical framework address this gap by providing:
1) \textit{Accountability}: Exposing critical vulnerabilities in deployed detectors to prioritize robustness over benchmark metrics.
2) \textit{Mitigation Pathways}: By quantifying failure modes (e.g., 30\% accuracy drops under occlusions), we enable targeted improvements in detection infrastructure.
3) \textit{Regulation Needs}: Our findings underscore the need for regulatory frameworks mandating detector testing against dynamic, real-world benchmarks before deployment.
Ultimately, this work empowers researchers to combat advanced deepfakes, fostering a more resilient digital social environment.

\noindent\textbf{Conclusion.}
This paper identifies and diagnoses the systemic failure of deepfake detectors in real-world scenarios through a multi-faceted study. We establish that two core limitations—temporal obsolescence and realism naivety —decouple laboratory performance from real-world efficacy. 
To bridge this gap, we introduce DeepFake-MIX, a comprehensive benchmark that contains:
1) \textit{Diverse AIGC Ecosystem}: 26 state-of-the-art forgery methods spanning face-swapping, synthesis, reenactment, and editing.
2) \textit{Real-World Perturbations}: Native social media artifacts (watermarks, danmu, filters) and simulated operations.
Our framework dissects detection vulnerabilities across critical factors:
1) \textit{Generation Models}: Regenerative face-swaps and audio-driven reenactment are hardest to detect.
2) \textit{Environmental Conditions}: Extreme illumination (e.g., top-heavy lighting) and head stances reduce accuracy by 15\%.
3) \textit{Social media protuberances}: Occlusions (danmu/filters) and retouching (heavy skin smoothing) degrade detection heavily.
Experiments with 8 SoTA detectors confirm alarming failures: Models plummet from 99\% (FF++) to 69\% on real-world data. 
Crucially, detectors trained on DeepFake-MIX show 10\% higher cross-domain robustness, validating its efficacy as a next-generation training foundation.

\bibliography{aaai2026}
\section{Reproducibility Checklist}
Unless specified otherwise, please answer ``yes'' to each question if the relevant information is described either in the paper itself or in a technical appendix with an explicit reference from the main paper. If you wish to explain an answer further, please do so in a section titled ``Reproducibility Checklist'' at the end of the technical appendix.

This paper:
\begin{itemize}
    \item Includes a conceptual outline and/or pseudocode description of AI methods introduced (yes)
    \item Clearly delineates statements that are opinions, hypothesis, and speculation from objective facts and results (yes)
    \item Provides well marked pedagogical references for less-familiare readers to gain background necessary to replicate the paper (yes)
\end{itemize}

\subsection{Theoretical Contributions}
Does this paper make theoretical contributions? (no)

If yes, please complete the list below.
\begin{itemize}
    \item All assumptions and restrictions are stated clearly and formally. (yes/partial/no)
    \item All novel claims are stated formally (e.g., in theorem statements). (yes/partial/no)
    \item Proofs of all novel claims are included. (yes/partial/no)
    \item Proof sketches or intuitions are given for complex and/or novel results. (yes/partial/no)
    \item Appropriate citations to theoretical tools used are given. (yes/partial/no)
    \item All theoretical claims are demonstrated empirically to hold. (yes/partial/no/NA)
    \item All experimental code used to eliminate or disprove claims is included. (yes/no/NA)
\end{itemize}

\subsection{Datasets}
Does this paper rely on one or more datasets? (yes)

If yes, please complete the list below.
\begin{itemize}
    \item A motivation is given for why the experiments are conducted on the selected datasets (yes)
    \item All novel datasets introduced in this paper are included in a data appendix. (yes)
    \item All novel datasets introduced in this paper will be made publicly available upon publication of the paper with a license that allows free usage for research purposes. (yes)
    \item All datasets drawn from the existing literature (potentially including authors' own previously published work) are accompanied by appropriate citations. (yes)
    \item All datasets drawn from the existing literature (potentially including authors' own previously published work) are publicly available. (yes)
    \item All datasets that are not publicly available are described in detail, with explanation why publicly available alternatives are not scientifically satisficing. (NA)
\end{itemize}

\subsection{Computational Experiments}
Does this paper include computational experiments? (yes)

If yes, please complete the list below.
\begin{itemize}
    \item This paper states the number and range of values tried per (hyper-) parameter during development of the paper, along with the criterion used for selecting the final parameter setting. (yes)
    \item Any code required for pre-processing data is included in the appendix. (yes).
    \item All source code required for conducting and analyzing the experiments is included in a code appendix. (yes)
    \item All source code required for conducting and analyzing the experiments will be made publicly available upon publication of the paper with a license that allows free usage for research purposes. (yes)
    \item All source code implementing new methods have comments detailing the implementation, with references to the paper where each step comes from (yes)
    \item If an algorithm depends on randomness, then the method used for setting seeds is described in a way sufficient to allow replication of results. (yes)
    \item This paper specifies the computing infrastructure used for running experiments (hardware and software), including GPU/CPU models; amount of memory; operating system; names and versions of relevant software libraries and frameworks. (yes)
    \item This paper formally describes evaluation metrics used and explains the motivation for choosing these metrics. (yes)
    \item This paper states the number of algorithm runs used to compute each reported result. (yes)
    \item Analysis of experiments goes beyond single-dimensional summaries of performance (e.g., average; median) to include measures of variation, confidence, or other distributional information. (yes)
    \item The significance of any improvement or decrease in performance is judged using appropriate statistical tests (e.g., Wilcoxon signed-rank). (yes)
    \item This paper lists all final (hyper-)parameters used for each model/algorithm in the paper's experiments. (yes)
\end{itemize}

\end{document}